\documentclass[sigconf]{acmart}
\AtBeginDocument{%
  \providecommand\BibTeX{{%
    \normalfont B\kern-0.5em{\scshape i\kern-0.25em b}\kern-0.8em\TeX}}}

\copyrightyear{2022} 
\acmYear{2022} 
\setcopyright{rightsretained} 
\acmConference[WWW '22 Companion]{Companion Proceedings of the Web Conference 2022}{April 25--29, 2022}{Virtual Event, Lyon, France}
\acmBooktitle{Companion Proceedings of the Web Conference 2022 (WWW '22 Companion), April 25--29, 2022, Virtual Event, Lyon, France}\acmDOI{10.1145/3487553.3524254}
\acmISBN{978-1-4503-9130-6/22/04}

\usepackage{multirow}
\usepackage{graphicx}

\begin{document}

\title{Supervised Contrastive Learning for Product Matching}

\author{Ralph Peeters}
\orcid{0000-0003-3174-2616}
\affiliation{%
  \institution{Data and Web Science Group}
  \institution{University of Mannheim}
  \city{Mannheim}
  \country{Germany}
}
\email{ralph@informatik.uni-mannheim.de}

\author{Christian Bizer}
\orcid{0000-0003-2367-0237}
\affiliation{%
  \institution{Data and Web Science Group}
  \institution{University of Mannheim}
  \city{Mannheim}
  \country{Germany}
}
\email{chris@informatik.uni-mannheim.de}

\renewcommand{\shortauthors}{Peeters and Bizer}

\begin{abstract}
  Contrastive learning has moved the state of the art for many tasks in computer vision and information retrieval in recent years. This poster is the first work that applies supervised contrastive learning to the task of product matching in e-commerce using product offers from different e-shops. More specifically, we employ a supervised contrastive learning technique to pre-train a Transformer encoder which is afterward fine-tuned for the matching task using pair-wise training data. We further propose a source-aware sampling strategy that enables contrastive learning to be applied for use cases in which the training data does not contain product identifiers. We show that applying supervised contrastive pre-training in combination with source-aware sampling significantly improves the state-of-the-art performance on several widely used benchmarks: For Abt-Buy, we reach an F1-score of 94.29 (+3.24 compared to the previous state-of-the-art), for Amazon-Google 79.28 (+ 3.7). For WDC Computers datasets, we reach improvements between +0.8 and +8.84 in F1-score depending on the training set size. Further experiments with data augmentation and self-supervised contrastive pre-training show that the former can be helpful for smaller training sets while the latter leads to a significant decline in performance due to inherent label noise. We thus conclude that contrastive pre-training has a high potential for product matching use cases in which explicit supervision is available.
\end{abstract}



\maketitle

\section{Introduction}
\label{sec:introduction}

Contrastive Learning is a form of deep learning with the goal of separating dissimilar instances while grouping similar instances together in the embedding space. The contrastive learning approach has seen success in the area of information retrieval~\cite{gaoSimCSESimpleContrastive2021} and computer vision~\cite{chenSimpleFrameworkContrastive2020,khoslaSupervisedContrastiveLearning2021}, where current approaches~\cite{khoslaSupervisedContrastiveLearning2021}  outperform methods that solely rely on cross-entropy-based learning. 

In this poster, we investigate the potential of contrastive learning for the problem of product matching in e-commerce. This task, a special case of entity matching, is usually defined as a binary pair-wise classification task, where two product offers from different sources are compared with each other and assigned the label match or non-match depending on whether they refer to the same real-world product or not. Successfully matching offers from multiple sources is a prerequisite for many e-commerce applications including price comparison portals and electronic marketplaces. 
Recent work~\cite{liDeepEntityMatching2020,peetersDualobjectiveFinetuningBERT2021} has shown that Transformer models are particularly well suited for product matching. We extend this work by investigating the usefulness of contrastive learning, more specifically contrastive pre-training of Transformer models for the task of product matching. We adopt a recent approach for supervised contrastive learning from computer vision called SupCon~\cite{khoslaSupervisedContrastiveLearning2021} for product matching tasks in which the training set contains product IDs, such as GTINs or MPNs. We further propose a source-aware sampling strategy that eliminates inter-source label-noise and enables contrastive pre-training to also be successfully applied to use cases without explicit identifiers. 
In summary, combining Transformer encoder networks and supervised contrastive learning, we achieve new state-of-the-art results on all tested benchmark datasets\footnote{\url{https://paperswithcode.com/task/entity-resolution/}}. The code for replicating our experiments is available on GitHub\footnote{\url{https://github.com/wbsg-uni-mannheim/contrastive-product-matching}}.

\section{Supervised Contrastive Learning for Product Matching}
\label{sec:supcon}

The SupCon contrastive loss~\cite{khoslaSupervisedContrastiveLearning2021} uses all examples in a batch to maximize the distances between an example and all of its negatives in that batch, as well as minimize the distances between an example and all of its positives. To achieve this, the method exploits label information about the training examples. 
All examples in a batch that do not carry the same label are treated as negatives. 
The original implementation of SupCon for computer vision uses a set of N randomly sampled example/label pairs (e.g. a picture of a dog and the label "dog") where each example is augmented twice using a random function from a set of augmentation functions, creating final input batches of 2N examples. Using the assigned labels, the contrastive loss among all examples of a batch can be calculated and is iteratively optimized using gradient descent optimization. 



Our approach for applying contrastive learning to product matching consists of two steps: (i) a contrastive pre-training step on batches of individual product offers using SupCon loss, followed by (ii) a fine-tuning step using matching and non-matching pairs of product offers. 
We use the RoBERTa base model as encoder architecture, which has been shown to achieve strong results across different product matching benchmark datasets as well as different training set sizes~\cite{peetersDualobjectiveFinetuningBERT2021}.

\subsection{Contrastive Pre-Training}
\label{subsec:contrastivestep}

\textbf{Labels for Contrastive Training:} Supervised contrastive training assumes that all examples that refer to the same entity share the same label, e.g. all pictures of dogs are labeled as "dog". The training sets of some product matching tasks include product identifiers, such as GTIN or MPN numbers, which can directly be used as labels for contrastive training. The training sets of other tasks do not provide explicit product identifiers but only label a certain amount of product offer pairs from different sources as matching or non-matching. For contrastive pre-training, we need to obtain explicit labels on the entity level, so that matching offers share the same label. To obtain such labels, we use the matching pairs from the training set and build a correspondence graph over all product offers, where the edges of the graph connect matching offers. We can then assign a unique label to each connected component of the graph so that matching offers share the same label.

\begin{figure}[htb]
  \centering
  \includegraphics[angle=90,width=0.8\linewidth]{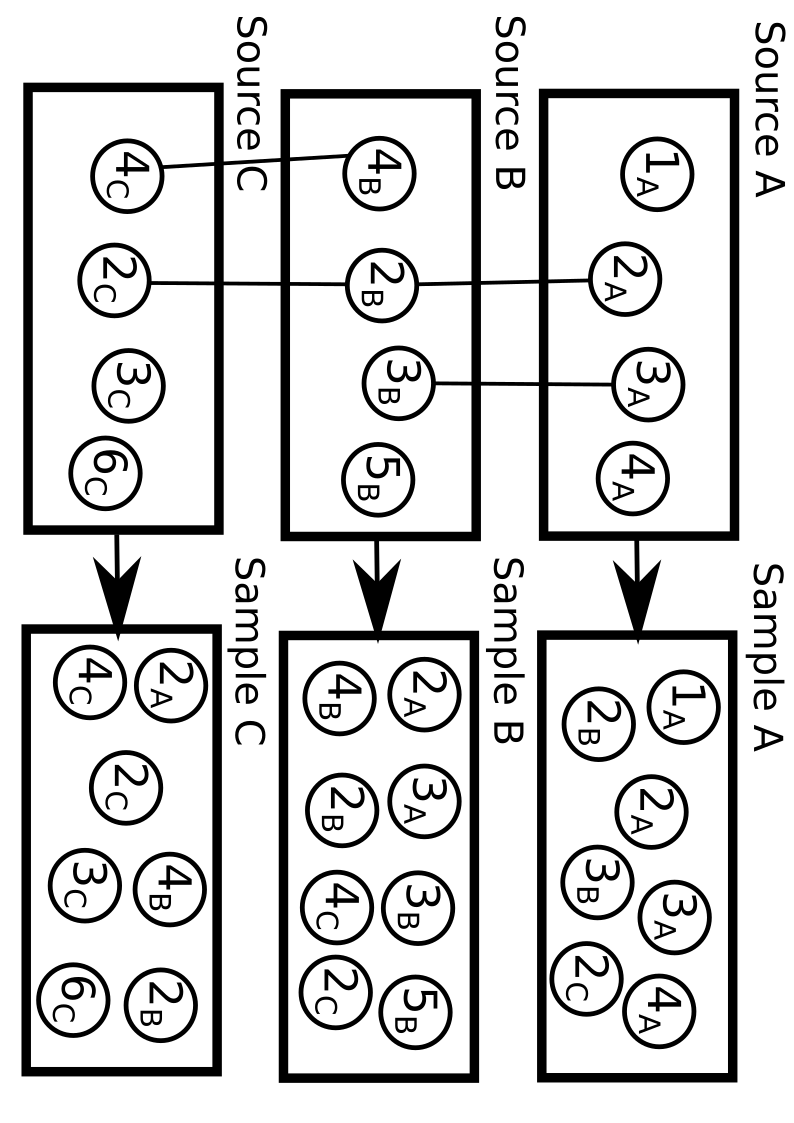}
  \caption{Sampling strategy for datasets with limited product identifier information. Produces one sampling dataset per source containing all product offers of that source as well as all known (connected) matching offers from other sources.}
  \Description{Sampling strategy for datasets with limited product identifier information. Produces one sampling dataset per source containing all product offers of that source as well as all known (connected) matching offers from other sources.}
  \label{fig:samplingstrat}
\end{figure}

\textbf{Source-aware Sampling Strategy:} Since we only know a subset of the matches between sources, some actually matching product offers will be assigned different labels during the previously presented process. During the contrastive pre-training step, this will then result in treating these as non-matching offers if they appear in the same batch due to the different label. This circumstance heavily deteriorates the quality of the learned representations as we discuss in Section \ref{subsec:results}. To alleviate this problem, we propose a source-aware sampling strategy that allows us to eliminate such inter-source label noise. Instead of generating one combined dataset containing all offers and their labels from each source, we instead generate one dataset per source, containing all offers from that source as well as only those offers from other sources which share a label with an offer from the current source, i.e. offers that were originally labeled as a match. Figure \ref{fig:samplingstrat} illustrates this procedure for a matching task involving three sources. 
Once a sampling dataset is built for every source using this procedure, we then sample offers from only one of the sampling datasets into each batch. For each batch, we randomly chose the dataset to sample from. This procedure allows us to completely eliminate inter-source label noise during contrastive training, given that the data sources themselves do not contain duplicates. 

\textbf{Batch Building Process:} We loosely follow the method from the original SupCon paper for assembling each batch. First, we select N product offers from a random sampling dataset and make use of available label information to randomly select for each of these offers a matching offer from the dataset. We allow selection of the same offer, even if other offers with the same label exist. The final batch then consists of 2N offers where for each of the N offers at least one offer having the same label is contained in the batch. All offers in a batch are then propagated through the encoder network to produce their vector representations, which are subsequently used to calculate the SupCon loss and tune the encoder parameters minimizing/maximizing distances between all matching/non-matching offers in the batch. 
As batches are sampled differently across epochs, many distance comparisons across all product offers are performed over all training epochs, leading to good representations in the learned vector space.

\textbf{Attribute Value Serialization:} 
As Transformer encoders expect inputs formatted as single strings, we serialize a product offer by concatenating all its attributes while still maintaining attribute separation by inserting additional tokens. More specifically, a single attribute is serialized as "[COL] column\_name [VAL] actual\_value". We then concatenate these strings for all attributes of a product offer to build its serialized input representation.

\textbf{Data Augmentation:} 
Applying data augmentations such as deleting words to product offers can easily distort an offer to an extent that the assigned label is no longer correct. For example, dropping the "4s" from the string "Apple iPhone 4s", would make it impossible to assign the correct label with certainty. Nevertheless, the usage of dropout noise during training inherent to Transformer encoders can be regarded as soft data augmentation in the embedding space, since two embeddings of the same product offer will likely never look exactly the same during training. 

In addition to only using the default dropout noise, we also experiment with applying explicit data augmentations to product offers during the pre-training stage. For this, we use the nlpaug\footnote{https://github.com/makcedward/nlpaug} python package and select 6 types of augmentations: (i) simulating typos, (ii) swapping words, (iii) deleting words, (iv) deleting spans of words, (v) substituting words with synonyms and (vi) randomly splitting words. For every selected offer in a batch, we randomly choose among all augmentations as well as the option of not augmenting the offer at all. If an offer is selected for augmentation, every word in that offer has a 10\% chance to be augmented with the currently selected augmentation method.

\subsection{Cross-Entropy Fine-Tuning}
\label{subsec:finetunestep}

For the fine-tuning step, we add a single dropout and linear layer on top of the model which returns a binary label, match or non-match, for a pair of product offers. We propagate both offers through the encoder and combine their mean-pooled representations as input to the final classification layer as follows: given the two embedding representations $u$ and $v$, we combine them as: $(u,v,|u-v|,u*v)$. The model is trained using binary cross-entropy loss. The parameters of the encoder layers can either be frozen or further tuned during the fine-tuning step, while the parameters of the classification layer are always tuned.

\section{Experiments}
\label{sec:experiments}
We evaluate the contrastive model on three benchmark datasets from the product matching domain: Abt-Buy, Amazon-Google, and WDC LSPC Computers. Abt-Buy and Amazon-Google represent the use-case of matching product offers from two deduplicated sources. The offers in both datasets do not contain product identifiers. 
The WDC LSPC dataset on the other hand represents a multi-source matching task. WDC LSPC training and validation set feature product identifiers for all offers. 

We implement our model using the huggingface Transformers library\footnote{https://github.com/huggingface/transformers}. For contrastive pre-training, the batch size is set to 1024. We leave the temperature at its default value of 0.07 and train using the Adam optimizer for 200 epochs with a linearly decaying learning rate of 5e-05 with a 0.05 warmup ratio. For the fine-tuning step, the batch size is set to 64 and we train for up to 50 epochs using early stopping if validation loss does not improve for 10 consecutive epochs. Each model is trained three times and we report the average results.

\subsection{Datasets}
\label{subsec:datasets}

\textbf{Abt-Buy/Amazon-Google:} For the two-source datasets Abt-Buy and Amazon-Google, we use the training, validation, and testing splits from the deepmatcher paper\footnote{https://github.com/anhaidgroup/deepmatcher/blob/master/Datasets.md}~\cite{mudgalDeepLearningEntity2018} and all available attributes, which allows us to directly compare the performance of our approach to recent matching systems using the same splits. Tables \ref{tab:training-sets} and \ref{tab:test-sets} shows statistics for both datasets. Abt-Buy and Amazon-Google do not contain product ids for single product offers in addition to the labeled offer pairs. To attain such identifiers, we apply the method described in Section \ref{subsec:contrastivestep} and subsequently apply our source-aware sampling strategy. The selection of pairs to use for building the correspondence graph is done only on the training and validation splits of the pairwise datasets. In an effort to introduce a regularizing effect, we further only use 80\% of matching pairs from the training and validation splits to perform this calculation. Due to the generally low amount of training offers per product in the two-source matching case, the model is more prone to overfit on the few known matching offers, reducing performance for products where no matching offers have been seen during contrastive pre-training. By withholding known matching information for 20\% of offer pairs, the model can later use these during the fine-tuning stage to better adapt to such cases.

\begin{table}[htb]
\centering
\caption{Training set statistics}
\label{tab:training-sets}
\resizebox{0.40\textwidth}{!}{%
\begin{tabular}{@{}lccccc@{}}
\toprule
Training Set &
  Size &
  \begin{tabular}[c]{@{}c@{}}\# Pos\\ Pairs\end{tabular} &
  \begin{tabular}[c]{@{}c@{}}\# Neg\\ Pairs\end{tabular} &
  \begin{tabular}[c]{@{}c@{}}\# offers for\\ pre-training\end{tabular} &
  \# products \\ \midrule
Abt-Buy                        & default & 822   & 6,837  & 2,112 & 1,084 \\
Amazon-Google                  & default & 933   & 8,234  & 3,445 & 2,279 \\ \midrule
\multirow{4}{*}{WDC computers} & xlarge  & 9,690 & 58,771 & 4,307 & 745   \\
                               & large   & 6,146 & 27,213 & 4,238 & 745   \\
                               & medium  & 1,762 & 6,332  & 3,846 & 745   \\
                               & small   & 722   & 2,112  & 2,790 & 745   \\ \bottomrule
\end{tabular}%
}
\end{table}

\begin{table}[htb]
\centering
\caption{Test set Statistics}
\label{tab:test-sets}
\resizebox{0.40\textwidth}{!}{%
\begin{tabular}{@{}lcccc@{}}
\toprule
Category &
  \begin{tabular}[c]{@{}c@{}}\# products\\ w/ pos (overall)\end{tabular} &
  \begin{tabular}[c]{@{}c@{}}\# Pos.\\ Pairs\end{tabular} &
  \begin{tabular}[c]{@{}c@{}}\# Neg.\\ Pairs\end{tabular} &
  \begin{tabular}[c]{@{}c@{}}\# Comb.\\ Pairs\end{tabular} \\ \midrule
Abt-Buy       & 205 (921) & 206 & 1,710 & 1,916 \\
Amazon-Google & 227 (1962) & 234 & 2,059 & 2,293 \\ \midrule
WDC computers & 150 (745) & 300 & 800   & 1,100 \\ \bottomrule
\end{tabular}%
}
\end{table}

\textbf{WDC LSCP:} We use the training, validation, and test sets from the computer's domain of WDC LSPC~\cite{primpeli2019wdc}. The training sets come in four different sizes, ranging from ~3K to ~70K product offer pairs. In addition to the pair labels, product offers are further annotated with product ids, which identify offers for the same products from different sources. We do not need to apply the source-aware sampling strategy and directly make use of these product ids as labels for contrastive pre-training and the pair-wise labels for the fine-tuning step. We use all product offers which are part of the training and validation sets for contrastive pre-training. The WDC LSPC dataset contains mainly textual attributes - we use the title, description, brand and specTableContent attributes. Tables \ref{tab:training-sets} and \ref{tab:test-sets} show the statistics of the WDC LSPC datasets.


\subsection{Results and Discussion}
\label{subsec:results}

\begin{table*}[htb]
\centering
\caption{F1-score results on the test sets for each dataset and training size. (F) and (UF) signify frozen and unfrozen encoder parameters during fine-tuning. For Abt-Buy and Amazon-Google results in brackets signify not reducing label-noise by separately sampling from both data sources. Results with * are taken from~\cite{liDeepEntityMatching2020}}
\label{tab:results}
\resizebox{0.7\textwidth}{!}{%
\begin{tabular}{@{}lcccccc@{}}
\toprule
\multicolumn{1}{c}{} &
  Abt-Buy &
  Amazon-Google &
  \multicolumn{4}{c}{WDC Computers} \\ \midrule
\begin{tabular}[c]{@{}l@{}}\# Training \\ Pairs\end{tabular} &
  $\sim$7.5K &
  $\sim$9K &
  \begin{tabular}[c]{@{}c@{}}$\sim$3K\\ (small)\end{tabular} &
  \begin{tabular}[c]{@{}c@{}}$\sim$8K\\ (medium)\end{tabular} &
  \begin{tabular}[c]{@{}c@{}}$\sim$23K\\ (large)\end{tabular} &
  \begin{tabular}[c]{@{}c@{}}$\sim$68K\\ (xlarge)\end{tabular} \\ \midrule
Deepmatcher &
  62.80* &
  70.70* &
  61.22 &
  69.85 &
  84.32 &
  88.95 \\
RoBERTa &
  91.05 &
  74.10* &
  86.37 &
  91.90 &
  94.68 &
  94.73 \\
Ditto &
  89.33* &
  75.58* &
  80.76* &
  88.62* &
  91.70* &
  95.45* \\
JointBERT &
  - &
  - &
  77.55 &
  88.82 &
  96.90 &
  97.49 \\ \midrule
R-SupCon(F) &
  93.70(38.24) &
  \textbf{79.28}(42.44) &
  93.18 &
  97.66 &
  98.16 &
  \textbf{98.33} \\
R-SupCon(F)+aug &
  \textbf{94.29} &
  76.14 &
  \textbf{95.21} &
  \textbf{98.50} &
  \textbf{98.50} &
  98.33 \\
R-SupCon(UF) &
  79.99(71.47) &
  71.81(61.06) &
  79.52 &
  87.32 &
  94.59 &
  96.16 \\
R-SupCon(UF)+aug &
  77.84 &
  68.37 &
  80.69 &
  89.12 &
  94.56 &
  96.13 \\ \midrule
R-SimCLR(F) &
  56.63 &
  56.16 &
  53.98 &
  55.25 &
  58.97 &
  60.66 \\
R-SimCLR(F)+aug &
  53.67 &
  54.29 &
  53.36 &
  54.97 &
  58.34 &
  62.19 \\
R-SimCLR(UF) &
  79.99 &
  64.87 &
  65.75 &
  82.72 &
  92.20 &
  95.25 \\
R-SimCLR(UF)+aug &
  79.28 &
  63.71 &
  66.73 &
  82.24 &
  91.89 &
  95.75 \\ \midrule
\begin{tabular}[c]{@{}l@{}}$\Delta$ to best baseline\end{tabular} &
  +3.24 &
  +3.7 &
  +8.84 &
  +6.6 &
  +1.6 &
  +0.84 \\ \bottomrule
\end{tabular}%
}
\end{table*}

We compare our results to recent neural entity matching systems: Ditto~\cite{liDeepEntityMatching2020}, JointBERT~\cite{peetersDualobjectiveFinetuningBERT2021}, Deepmatcher~\cite{mudgalDeepLearningEntity2018} and RoBERTa~\cite{liu_roberta_2019}. Since all of these systems were evaluated using the same train, validation, and test splits, we directly report the results from the corresponding papers. We evaluate two versions of contrastively pre-trained RoBERTa models: (i) R-SupCon using the supervised SupCon loss, (ii) R-SimCLR using self-supervision only, corresponding to SimCLR loss~\cite{chenSimpleFrameworkContrastive2020}. For R-SimCLR, each product offer is assigned a unique id and a match for each offer is only sampled by augmenting the same offer either via implicit dropout noise or explicit data augmentation. Table 3 shows the results of applying contrastive pre-training in comparison to the four baseline systems.

For the two-source datasets Abt-Buy and Amazon-Google, applying contrastive pre-training results in an improvement of 3.2-3.7\% F1 compared to the respective strongest baseline model. As stated in Section \ref{subsec:datasets}, we report results for two sampling strategies for contrastive learning, one containing all offers from all sources in the sampling set leading to clear label-noise, and the other using our source-aware sampling strategy with separated sampling datasets to eliminate such noise. The experiments clearly show that it is important to reduce label-noise for supervised contrastive learning for such cases: For Abt-Buy and Amazon-Google performance drops by 55\% and 37\% F1 respectively without the source-aware sampling strategy. All contrastively pre-trained RoBERTa models outperform the baselines by 0.8-8.8\% F1 for the WDC dataset. Adding contrastive pre-training can improve on the best baseline results for small and medium training sizes by 8.8\% and 6.6\% F1 respectively. Improvements on large and xlarge are visible but comparably small in the range of 0.8\% to 1.8\% F1.  In general, freezing the encoder parameters after the contrastive pre-training step leads to higher performance on all datasets compared to further updating them during the fine-tuning step. 

Applying augmentation during the contrastive pre-training phase delivers mixed results across datasets. For the smaller WDC training sets, we see an improvement of 1-2\% F1 but only minimal improvements on the larger training sets and on Abt-Buy, while Amazon-Google sees a 4\% drop in performance when applying augmentation.

Using only self-supervision during contrastive pre-training leads to worse results compared to not pre-training at all. The label-noise inherent to self-supervised SimCLR due to only regarding an augmented version of the same offer as matching and treating all others as non-matching, even if they do actually match, is likely the cause of the large drop in performance.


\section{Conclusion}
\label{sec:conclusion}

We have demonstrated that supervised contrastive pre-training followed by cross-entropy fine-tuning can generally improve the performance of product matchers compared to only performing cross-entropy fine-tuning on multi-source as well as two-source benchmark tasks by 0.8-8.8\% in F1-score, thereby setting a new state-of-the-art for each of the tasks. 
We further proposed a source-aware sampling strategy designed to reduce inter-source label noise during contrastive pre-training. We showed that this sampling procedure is crucial for achieving good performance on matching tasks without explicit product identifiers. Performing the contrastive pre-training step in a purely self-supervised fashion decreases the performance of the fine-tuned matchers likely due to the same effect. 
In summary, we show that supervised contrastive pre-training is a promising technique for product matching. 

\begin{acks}
The authors acknowledge support by the state of Baden-Württemberg through bwHPC.
\end{acks}

\bibliographystyle{ACM-Reference-Format}
\bibliography{main}

\end{document}